\documentclass[english,aip, jcp, reprint]{revtex4-1}
\usepackage[T1]{fontenc}
\usepackage[latin9]{inputenc}
\usepackage{amsbsy}
\usepackage{amstext}
\usepackage{amssymb}
\usepackage{graphicx}
\usepackage{babel}
\begin{document}

\title{Time-lagged autoencoders: \\
Deep learning of slow collective variables for molecular kinetics}

\author{Christoph Wehmeyer}
\email{christoph.wehmeyer@fu-berlin.de}

\selectlanguage{english}%

\author{Frank Noé}
\email{frank.noe@fu-berlin.de}

\selectlanguage{english}%

\affiliation{Freie Universität Berlin, Department of Mathematics and Computer
Science, Arnimallee 6, 14195 Berlin, Germany}
\begin{abstract}
Inspired by the success of deep learning techniques in the physical
and chemical sciences, we apply a modification of an autoencoder type
deep neural network to the task of dimension reduction of molecular
dynamics data. We can show that our time-lagged autoencoder reliably
finds low-dimensional embeddings for high-dimensional feature spaces
which capture the slow dynamics of the underlying stochastic processes\textendash beyond
the capabilities of linear dimension reduction techniques.
\end{abstract}
\maketitle
Molecular dynamics (MD) simulation allows us to probe the full spatiotemporal
detail of molecular processes, but its usefulness has long been limited
by the sampling problem. Recently, the combination of hard- and software
for high-throughput MD simulations \citep{ShirtsPande_Science2000_FoldingAtHome,BuchEtAl_JCIM10_GPUgrid,Shaw_Science10_Anton,PronkEtAl_Bioinf13_Gromacs4.5,DoerrEtAl_JCTC16_HTMD}
with Markov state models (MSM) \citep{msm-jhp,msm-book,SarichSchuette_MSMBook13}
has enabled the exhaustive statistical description of protein folding
\citep{NoeSchuetteReichWeikl_PNAS09_TPT,Bowman_JCP09_Villin,LindorffLarsenEtAl_Science11_AntonFolding},
conformational changes \citep{SadiqNoeFabritiis_PNAS12_HIV,KohlhoffEtAl_NatChem14_GPCR-MSM},
protein-ligand association \citep{BuchFabritiis_PNAS11_Binding,SilvaHuang_PlosCB_LaoBinding,PlattnerNoe_NatComm15_TrypsinPlasticity}
and even protein-protein association \citep{PlattnerEtAl_NatChem17_BarBar}.
Using multi-ensemble Markov models (MEMMs) \citep{WuMeyRostaNoe_JCP14_dTRAM,RostaHummer_DHAM,WuEtAL_PNAS16_TRAM,MeyWuNoe_xTRAM},
even the kinetics of ultra-rare events beyond the seconds timescale
are now available at atomistic resolution \citep{trammbar,CasasnovasEtAl_JACS17_UnbindingKinetics,TiwaryEtAl_SciAdv17_DrugUnbinding}.
A critical step in Markov state modeling and MSM-based sampling is
the drastic dimension reduction from molecular configuration space
to a space containing the slow collective variables (CVs) \citep{NoeClementi_COSB17_SlowCVs,PretoClementi_PCCP14_AdaptiveSampling,McCartyParrinello_VACMetadynamics}.

Another area that has made recent breakthroughs is deep learning,
with impressive success in a variety of applications \citep{deep-learning,alphago}
that indicate the capabilities of deep neural networks to uncover
hidden structures in complex datasets. More recently, machine learning
has also been successfully applied to chemical physics problems, such
as learning quantum-chemical potentials, data-driven molecular design
and binding site prediction \citep{mlmae,rnn-ae-smiles,Goh-dl4cc,qmdtnn,tuckerman-ann-cv,JimenezDeFabritiis_Bioinf17_Deepsite}.
In the present study, we demonstrate that a deep time-lagged autoencoder
network \citep{autoencoder-origins,Baldi-autoencoder} can be employed
to perform the drastic dimension reduction required and find slow
CVs that are suitable to build highly accurate MSMs.

Identification of slow CVs, sometimes called reaction coordinates,
is an active area of research \citep{PetersTrout_JCP06_ReactionCoordinateOptimization,RohrdanzClementi_JCP134_DiffMaps,RohrdanzEtAl_AnnRevPhysChem13_MountainPasses,StamatiEtAl_Proteins10_NonlinearDimRed,DasEtAl_PNAS06_NonlinearDimred,Peters_jcp06_committorerrorestimation,PrinzChoderaNoe_PRX14_RateTheory,AltisStock_JCP07_dPCA,KrivovKarplus_PNAS101_14766}.
State of the art methods for identifying slow CVs for MD are based
on the variational approach for conformation dynamics (VAC) \citep{NoeNueske_MMS13_VariationalApproach,NueskeEtAl_JCTC14_Variational}
and its recent extension to non-equilibrium processes \citep{WuNoe_VAMP},
which provide a systematic framework for finding the optimal slow
CVs for a given time series. A direct consequence of the VAC is that
time-lagged independent component analysis (TICA) \citep{tica,tica3},
originally developed for blind-source separation \citep{tica2,ZieheMueller_ICANN98_TDSEP},
approximate the optimal slow CVs by linear combinations of molecular
coordinates. An very similar method, developed in the dynamical systems
community is dynamic mode decomposition (DMD) \citep{Mezic_NonlinDyn05_Koopman,SchmidSesterhenn_APS08_DMD,TuEtAl_JCD14_ExactDMD}.
VAC and DMD find slow CVs \emph{via} two different optimization goals
using the time series $\{\mathbf{z}_{t}\}$:
\begin{enumerate}
\item \emph{Variational approach} (VAC): search the $d$ orthogonal directions
$\mathbf{r}_{i}$, $i=1,...,d$, such that the time-lagged autocorrelation
of the projection $\mathbf{r}_{i}^{\top}\mathbf{z}_{t}$ is maximal
\citep{tica2}. These autocorrelations are bounded from above by the
eigenvalues of the Markov propagator \citep{NoeNueske_MMS13_VariationalApproach}.
\item \emph{Regression approach} (DMD): find the linear model $\mathbf{K}$
with the minimal regression error $\sum_{t}\left\Vert \mathbf{z}_{t+\tau}-\mathbf{K}^{\top}\mathbf{z}_{t}\right\Vert ^{2}$
and compute its $d$ eigenvectors $\mathbf{r}_{i}$ with largest eigenvalues.
\end{enumerate}
Both approaches will give us the same directions $\mathbf{r}_{i}$
if the corresponding sets of eigenvectors are used \citep{KlusEtAl_JNS17_DataDriven}.
These directions can be used for dimension reduction. By experience,
we know that the dimension reduction can be made much more efficient
by working in feature space instead of directly using the Cartesian
coordinates \citep{HarmelingEtAl_NeurComput03_KernelTDSEP,PetersTrout_JCP06_ReactionCoordinateOptimization,DasEtAl_PNAS06_NonlinearDimred,RohrdanzClementi_JCP134_DiffMaps,NoeNueske_MMS13_VariationalApproach,SchwantesPande_JCTC15_kTICA,NueskeEtAl_JCTC14_Variational,WilliamsEtAl_Arxiv14_KernelEDMD,NueskeEtAl_JCP15_Tensor,BruntonProctorKutz_PNAS16_Sindy,HarriganPande_bioRxiv17_LandmarkTICA}.
That means we perform some nonlinear mapping:
\begin{equation}
\mathbf{e}_{t}=E(\mathbf{z}_{t}),\label{eq:feature_mapping}
\end{equation}
e.g., by computing distances between residues or torsion angles, and
then perform TICA or DMD (then known as EDMD \citep{WilliamsKevrekidisRowley_JNS15_EDMD})
in the $\mathbf{e}_{t}$ coordinates. Indeed this approach is fully
described by the VAC \citep{NoeNueske_MMS13_VariationalApproach}
and will provide an optimal approximation of the slow components of
the dynamics \emph{via} a linear combination of the feature functions.
If we do not want to choose the library of feature functions by hand,
but instead want to optimize the nonlinear mapping $E$ by employing
a neural network, we have again two options: (1) employ the variational
approach. This leads to VAMPnets described in \citep{MardtEtAl_VAMPnets},
or (2) minimize the regression error:
\begin{equation}
\min_{D,E}\sum_{t}\left\Vert \mathbf{z}_{t+\tau}-D(E(\mathbf{z}_{t}))\right\Vert ^{2},\label{eq:TAE_regression_error}
\end{equation}
where $D$ is some mapping from feature space to coordinate space
and also includes the time-propagation. In this paper we investigate
option (2), which naturally leads to using a time-lagged autoencoder
(TAE).

\begin{figure}
\includegraphics{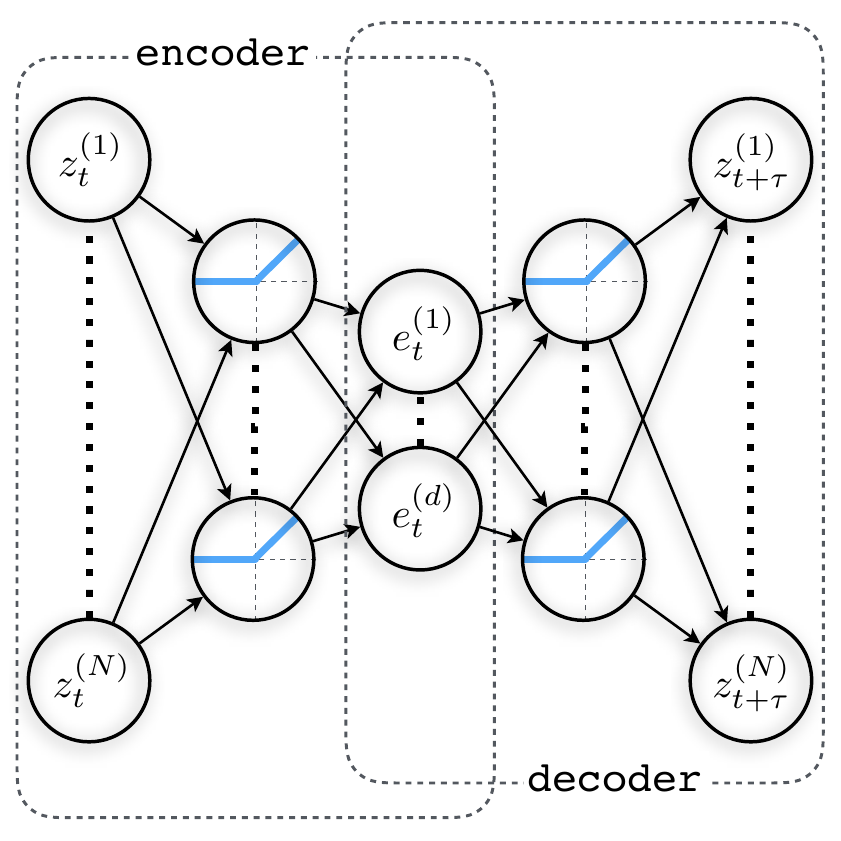}

\caption{\label{fig:autoencoder}Schematic of a time-lagged autoencoder type
neural network with one nonlinear hidden layer in each the encoding
and decoding part. The encoder transforms a vector $\mathbf{z}_{t}\in\mathbb{R}^{N}$
to a $d$-dimensional latent space while the decoder maps this latent
vector $\text{\ensuremath{\mathbf{e}}}_{t}$ to the vector $\mathbf{z}_{t+\tau}\in\mathbb{R}^{N}$
in full coordinate space but at time $\tau$ later. For $\tau=0$,
this setup corresponds to a regular autoencoder.}
\end{figure}

An autoencoder (Fig. \ref{fig:autoencoder}), is a type of deep neural
network which is trained in a self-supervised manner \citep{autoencoder-origins,Baldi-autoencoder}.
The layer structure of the network is usually symmetric with a bottleneck
in the middle and we refer to the first half including the bottleneck
as the encoder while the second half is called decoder. Such a network
is then trained to reconstruct its actual input $\mathbf{z}_{t}$
with minimal regression error, i.e., the network must learn to encode
an $N$-dimensional vector as a $d$-dimensional representation to
pass the information through the bottleneck and reconstruct the original
signal again in the decoder. Autoencoders can be viewed as a nonlinear
version of a rank-$d$ principal component analysis (PCA), and one
can show that a linear autoencoder with bottleneck size $d$ will
identify the space of the $d$ largest principal components \citep{BaldiHornik_JNN89_NNPCA}.
Autoencoders have been successfully applied to de-noise images \citep{ann-dim-redux,Wang_2016}
and reduce the dimensionality of molecular conformations \citep{Brown2008}.

In this study, we alter the self-supervised training in such a way
that the network minimizes the DMD regression error defined in Eq.
(\ref{eq:TAE_regression_error}), i.e., instead of training the network
to reconstruct its input ($\mathbf{z}_{t}\mapsto\mathbf{z}_{t}$),
we train it to predict a later frame ($\mathbf{z}_{t}\mapsto\mathbf{z}_{t+\tau}$);
this is still self-supervised but requires to train on time series.
While there are very strong mathematical arguments for employing the
variational approach to learn nonlinear feature transformations \emph{via}
VAMPnets (see discussion in \citep{WuNoe_VAMP,MardtEtAl_VAMPnets}),
there are practical arguments why the regression approach taken in
the present TAEs is attractive: (i) we do not only learn the encoder
network that performs the dimension reduction, but we also learn the
decoder network that can predict samples in our original coordinate
space from points in the latent space, and (ii) TAEs can be extended
towards powerful sampling methods such as variational and adversarial
autoencoders \citep{KingmaWelling_ICLR14_VAE,MakhzaniEtAl_Arxiv16_AdversarialAutoencoders}.
Here, we demonstrate that deep TAEs perform equally well or better
than state of the art methods for finding the slow CVs in stochastic
dynamical systems and biomolecules.

\section{Theory}

To motivate our approach we first consider the case of linear transformations.
We show that in a similar way that a linear autoencoder and PCA are
equivalent up to orthogonalization, linear TAEs are equivalent to
time-lagged canonical correlation analysis (TCCA), and in the time-reversible
case equivalent to TICA. Then we move on to employ nonlinear TAEs
for dimension reduction.

We are given a time-series with $N$ dimensions and $T$ time-steps,
$\lbrace\mathbf{z_{t}}\in\mathbb{R}^{N}\rbrace_{t=1}^{T}$, and search
for a $d$-dimensional embedding ($d<N$) which is well suited to
compress the time-lagged data. To this aim, we define an encoding
$E:\mathbb{R}^{N}\to\mathbb{R}^{d}$ and a decoding $D:\mathbb{R}^{d}\to\mathbb{R}^{N}$
operation which approximately reconstruct the time-lagged signal such
that, on average, the error
\[
\boldsymbol{\epsilon}_{t}=\mathbf{z}_{t+\tau}-D(E(\mathbf{z}_{t}))
\]
is small in some suitable norm. We introduce two conventions:
\begin{enumerate}
\item we employ mean-free coordinates,
\begin{eqnarray*}
\mathbf{x}_{t} & = & \mathbf{z}_{t}-\frac{1}{T-\tau}\sum_{s=1}^{T-\tau}\mathbf{z}_{s}\\
\mathbf{y}_{t} & = & \mathbf{z}_{t+\tau}-\frac{1}{T-\tau}\sum_{s=1}^{T-\tau}\mathbf{z}_{s+\tau},
\end{eqnarray*}
\item and whiten them,
\begin{eqnarray*}
\tilde{\mathbf{x}}_{t} & = & \mathbf{C}_{00}^{-\frac{1}{2}}\mathbf{x}_{t}\\
\tilde{\mathbf{y}}_{t} & = & \mathbf{C}_{\tau\tau}^{-\frac{1}{2}}\mathbf{y}_{t},
\end{eqnarray*}
using the covariance matrices
\begin{eqnarray}
\mathbf{C}_{00} & = & \frac{1}{T-\tau}\sum_{t=1}^{T-\tau}\mathbf{x}_{t}\mathbf{x}_{t}^{\top}\label{eq:C00}\\
\mathbf{C}_{0\tau} & = & \frac{1}{T-\tau}\sum_{t=1}^{T-\tau}\mathbf{x}_{t}\mathbf{y}_{t}^{\top}\label{eq:C0t}\\
\mathbf{C}_{\tau\tau} & = & \frac{1}{T-\tau}\sum_{t=1}^{T-\tau}\mathbf{y}_{t}\mathbf{y}_{t}^{\top}.\label{eq:Ctt}
\end{eqnarray}
Note that whitening must take into account that $\mathbf{C}_{00}$
and $\mathbf{C}_{\tau\tau}$ are often not full rank matrices \citep{WuEtAl_JCP17_VariationalKoopman}.
\end{enumerate}
Now we must find an encoding and decoding which minimizes the reconstruction
error
\begin{equation}
\min_{E,D}\sum_{t=1}^{T-\tau}\|\tilde{\mathbf{y}}_{t}-D(E(\tilde{\mathbf{x}}_{t}))\|_{2}^{2}\label{eq:minimal_error}
\end{equation}
for a selected class of functions $E$ and $D$. The simplest choice,
of course, are linear functions.

\subsection{Linear TAE performs TCCA}

As we operate on mean free data, we can represent linear encodings
and decodings by simple matrix multiplications:
\begin{eqnarray*}
E(\tilde{\mathbf{x}}_{t}) & = & \tilde{\mathbf{E}}\tilde{\mathbf{x}}_{t}\\
D(E(\tilde{\mathbf{x}}_{t})) & = & \tilde{\mathbf{D}}\tilde{\mathbf{E}}\tilde{\mathbf{x}}_{t}
\end{eqnarray*}
with the encoding matrix $\tilde{\mathbf{E}}\in\mathbb{R}^{d\times N}$,
which projects $N$-dimensional data onto a $d$-dimensional space,
and the decoding matrix $\tilde{\mathbf{D}}\in\mathbb{R}^{N\times d}$,
which lifts the encoded data back to an $N$-dimensional vector space.

For convenience, we define the matrices $\mathbf{X}=\left[\mathbf{\mathbf{x}_{1},\dots,}\mathbf{x}_{T-\tau}\right]$
and $\mathbf{Y}=\left[\mathbf{\mathbf{y}_{1},\dots,}\mathbf{y}_{T-\tau}\right]$,
and likewise $\tilde{\mathbf{X}},\tilde{\mathbf{Y}}$ for the whitened
coordinates. The minimal reconstruction error (\ref{eq:minimal_error})
thus becomes
\begin{equation}
\min_{\tilde{\mathbf{K}}_{d}}\|\tilde{\mathbf{Y}}-\tilde{\mathbf{K}}_{d}\tilde{\mathbf{X}}\|_{F}^{2},\label{eq:minimal_error_linear}
\end{equation}
where $F$ denotes the Frobenius norm and $\tilde{\mathbf{K}}_{d}=\tilde{\mathbf{D}}\tilde{\mathbf{E}}$
is the rank-$d$ Koopman matrix; for the full-rank Koopman matrix,
we simply omit the rank index: $\tilde{\mathbf{K}}_{N}=\tilde{\mathbf{K}}$.

Eq. (\ref{eq:minimal_error_linear}) is a linear least squares problem.
The full-rank solution is given by the regression
\[
\tilde{\mathbf{K}}=\tilde{\mathbf{Y}}\tilde{\mathbf{X}}^{\top}\left(\tilde{\mathbf{X}}\tilde{\mathbf{X}}^{\top}\right)^{-1}=\frac{1}{T-\tau}\tilde{\mathbf{Y}}\tilde{\mathbf{X}}^{\top},
\]
where we have used the fact that the data is whitened: $\tilde{\mathbf{X}}\tilde{\mathbf{X}}^{\top}=(T-\tau)\mathbf{I}$.
Using the definition of the covariance matrices (\ref{eq:C00}-\ref{eq:Ctt}),
we can also write
\[
\tilde{\mathbf{K}}^{\top}=\tilde{\mathbf{X}}\tilde{\mathbf{Y}}^{\top}=\mathbf{C}_{00}^{-\frac{1}{2}}\mathbf{C}_{0\tau}\mathbf{C}_{\tau\tau}^{-\frac{1}{2}}.
\]
This form is often referred to as half-weighted Koopman matrix and
it arises naturally when using whitened data.

The last step to the solution lies in choosing the optimal rank-$d$
approximation to the Koopman matrix which is given by the rank-$d$
singular value decomposition,
\begin{eqnarray*}
\tilde{\mathbf{K}}_{d}^{\top} & = & \text{svd}_{d}\left(\mathbf{C}_{00}^{-\frac{1}{2}}\mathbf{C}_{0\tau}\mathbf{C}_{\tau\tau}^{-\frac{1}{2}}\right)\\
 & = & \mathbf{U}_{d}\mathbf{\Sigma}_{d}\mathbf{V}_{d}^{\top},
\end{eqnarray*}
where $d$ indicates that we take the $d$ largest singular values
and corresponding singular vectors. Thus, a possible choice for the
encoding and decoding matrices for whitened data is
\begin{eqnarray*}
\tilde{\mathbf{E}} & = & \mathbf{\Sigma}_{d}\mathbf{V}_{d}^{\top}\\
\tilde{\mathbf{D}} & = & \mathbf{U}_{d}.
\end{eqnarray*}
With this choice, we find for the mean-free but non-whitened data
\begin{eqnarray*}
\tilde{\mathbf{y}}_{t} & = & \tilde{\mathbf{K}}\tilde{\mathbf{x}}_{t}\\
\Leftrightarrow\mathbf{C}_{\tau\tau}^{-\frac{1}{2}}\mathbf{y}_{t} & = & \left(\mathbf{C}_{00}^{-\frac{1}{2}}\mathbf{C}_{0\tau}\mathbf{C}_{\tau\tau}^{-\frac{1}{2}}\right)^{\top}\mathbf{C}_{00}^{-\frac{1}{2}}\mathbf{x}_{t}\\
\Leftrightarrow\mathbf{y}_{t} & = & \mathbf{C}_{0\tau}^{\top}\mathbf{C}_{00}^{-1}\mathbf{x}_{t}
\end{eqnarray*}
the non-whitened Koopman matrix consistently with \citep{NoeNueske_MMS13_VariationalApproach,WilliamsKevrekidisRowley_JNS15_EDMD,WuNoe_VAMP,HorenkoHartmannSchuetteNoe_PRE07_Langevin}:
\[
\mathbf{K}^{\top}=\mathbf{C}_{00}^{-1}\mathbf{C}_{0\tau}.
\]
Likewise, we find the non-whitened encoding and decoding matrices
\begin{eqnarray*}
\mathbf{E} & = & \mathbf{\Sigma}_{d}\mathbf{V}_{d}^{\top}\mathbf{C}_{00}^{-\frac{1}{2}}\\
\mathbf{D} & = & \mathbf{C}_{\tau\tau}^{\frac{1}{2}}\mathbf{U}_{d},
\end{eqnarray*}
where the encoding consists of a whitening followed by the whitened
encoding, while the decoding starts with the whitened decoding followed
by unwhitening. This solution is equivalent with time-lagged canonical
correlation analysis (TCCA) \citep{Hotelling_Biometrika36_CCA,WuNoe_VAMP}.

\subsection{Time-reversible linear TAE performs TICA}

If the covariance matrix $C_{0\tau}$ is symmetric, the singular value
decomposition of the full-rank Koopman matrix is equivalent to an
eigenvector decomposition:
\[
\tilde{\mathbf{K}}_{d}=\mathbf{U}_{d}\mathbf{\Sigma}_{d}\mathbf{U}_{d}^{\top}.
\]
If we further have a stationary time series, i.e., $\mathbf{C}_{00}=\mathbf{C}_{\tau\tau}$,
the non-whitened encoding and decoding matrices are
\begin{eqnarray*}
\mathbf{E} & = & \mathbf{\Sigma}_{d}\mathbf{U}_{d}^{\top}\mathbf{C}_{00}^{-\frac{1}{2}}\\
\mathbf{D} & = & \mathbf{C}_{00}^{\frac{1}{2}}\mathbf{U}_{d},
\end{eqnarray*}
where $\mathbf{U}_{d}^{\top}\mathbf{C}_{00}^{-\frac{1}{2}}$ contains
the usual TICA eigenvectors and multiplication with $\mathbf{\Sigma}_{d}$
transforms to a kinetic map. Thus, if we include $\mathbf{\Sigma}_{d}$
in the decoder part, this solution is equivalent to TICA \citep{tica2,tica,tica3},
while if we include it in the encoder, it is equivalent to a kinetic
map \citep{kinetic-maps}.

Motivated by these theoretical results we will employ TAEs to learn
nonlinear encodings and decodings that optimize Eq. (\ref{eq:minimal_error}).

\section{Experiments}

We put the nonlinear time-lagged autoencoder to the test by applying
it to two toy models of different degree of difficulty as well as
molecular dynamics data for alanine dipeptide. In all three cases,
we compare the performance of the autoencoder with that of TICA (with
kinetic map scaling) and PCA by
\begin{enumerate}
\item comparing the reconstruction errors (\ref{eq:minimal_error}) of the
validation sets,
\item comparing the low-dimensional representations found with the known
essential variables of the respective system by employing canonical
correlation analysis (CCA), and 
\item examining the suitability of the encoded space for building MSMs via
convergence of implied timescales.
\end{enumerate}
The time-lagged autoencoders used in this study are implemented using
the PyTorch framework \citep{pytorch-repo} and consist of an input
layer with $N$ units, followed by one or two hidden layers with sizes
$H_{1}$ and $H_{2}$ and the latent layer with size $d$ which concludes
the encoding stage. The decoding part also adds one or two hidden
layers of the same sizes as in the encoding part, followed by the
output layer with size $N$. All hidden layers employ leaky rectified
linear units \citep{relu} (leaky parameter $\alpha=0.001$) and a
dropout layer \citep{dropout} (dropout probability $p=0.5$). We
train the networks using the Adam \citep{adam-optimizer} optimizer.

To account for the stochastic nature of the high dimensional data,
the autoencoder training process, and the discretization when building
MSMs, all simulations have been repeated 100 times while shuffling
training and validation sets. We show the ensemble median as well
as a one-standard-deviation percentile (68\%). The evaluation process
always follows the pattern
\begin{enumerate}
\item Gather the high dimensional data and reference low-dimensional representation
\emph{via} independent simulation or bootstrapping.
\item Train the encoder/decoder for all techniques on two thirds (training
set) of the high dimensional data.
\item Compute the reconstruction error for the remaining third of the data
(validation set).
\item Obtain encoded coordinates and whiten (training + validation sets).
\item Perform CCA to compare the encoded space to the reference data (training
+ validation sets).
\item Build MSMs \citep{pyemma} on the encoded space (training + validation
sets) and validate using implied timescales tests \citep{SwopePiteraSuits_JPCB108_6571}.
\end{enumerate}

\subsection{Two-state toy model}

\begin{figure}
\includegraphics{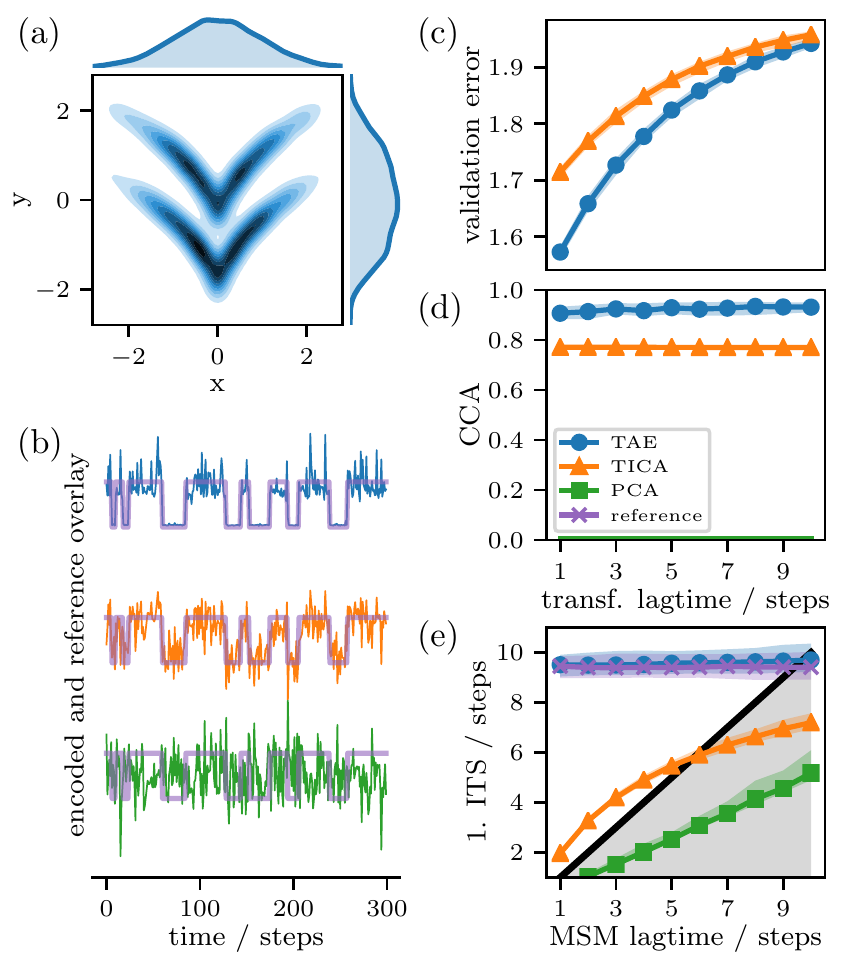}

\caption{\label{fig:sqrt}\textbf{Two-dimensional two-state system that is
not linearly separable}. (a) Joint and marginal distributions of the
two-state HMM toy model. (b) Comparison of time-series segments of
one-dimensional transformations (lag time $\tau=1$ step) with the
actual hidden state time series. (c) Regression error for the validation
set. (d) Canonical correlation coefficient between encoded time series
and true hidden state timeseries as a function of the transformation
lagtime. (e) Convergence of the slowest implied timescale for the
one-dimensional transformations and the hidden state time series.
Panels (c-e) show the median (lines) and 68\% percentiles (shaded
areas) over 100 independent realizations.}
\end{figure}

The first toy model is based on a two-state hidden Markov model (HMM)
which emits anisotropic Gaussian noise in the two-dimensional $x/y$-plane.
To complicate matters we perform the operation
\[
\left(\begin{array}{c}
x\\
y
\end{array}\right)\mapsto\left(\begin{array}{c}
x\\
y+\sqrt{x}
\end{array}\right)
\]
which leads to the distribution shown in Fig. \ref{fig:sqrt}a. We
compare one-dimensional representations found by applying TICA and
PCA to the time series, and a TAE employing one hidden layer of 50
units in the encoding and decoding part each, and a bottleneck size
of $d=1$.

The TAE-encoded variable overlaps very well with the hidden state
time series and can clearly separate both hidden states, while TICA
gives a more blurred picture with no clear separation and PCA does
not seem to separate the hidden states at all (Fig. \ref{fig:sqrt}b).
These differences are quantified by the CCA score between the encoded
and true hidden state signals (Fig. \ref{fig:sqrt}d). The time-lagged
autoencoder outperforms TICA at all examined transformation lagtimes
in terms of the reconstruction error; the difference is particularly
strong for small lagtimes (Fig. \ref{fig:sqrt}c). Finally, the encoding
found by the TAE is excellently suited to build an MSM that approximates
the slowest relaxation timescale even at short lagtimes (Fig. \ref{fig:sqrt}e).
In contrast, the MSM based on TICA converges towards the true timescale
too slowly, and does not get close to it before reaching the numerically
invalid range $\tau>t_{2}$. The MSM build on PCA seems to be completely
unsuitable for recovering kinetics (Fig. \ref{fig:sqrt}e).

\subsection{Four-state swissroll toy model}

\begin{figure}
\includegraphics{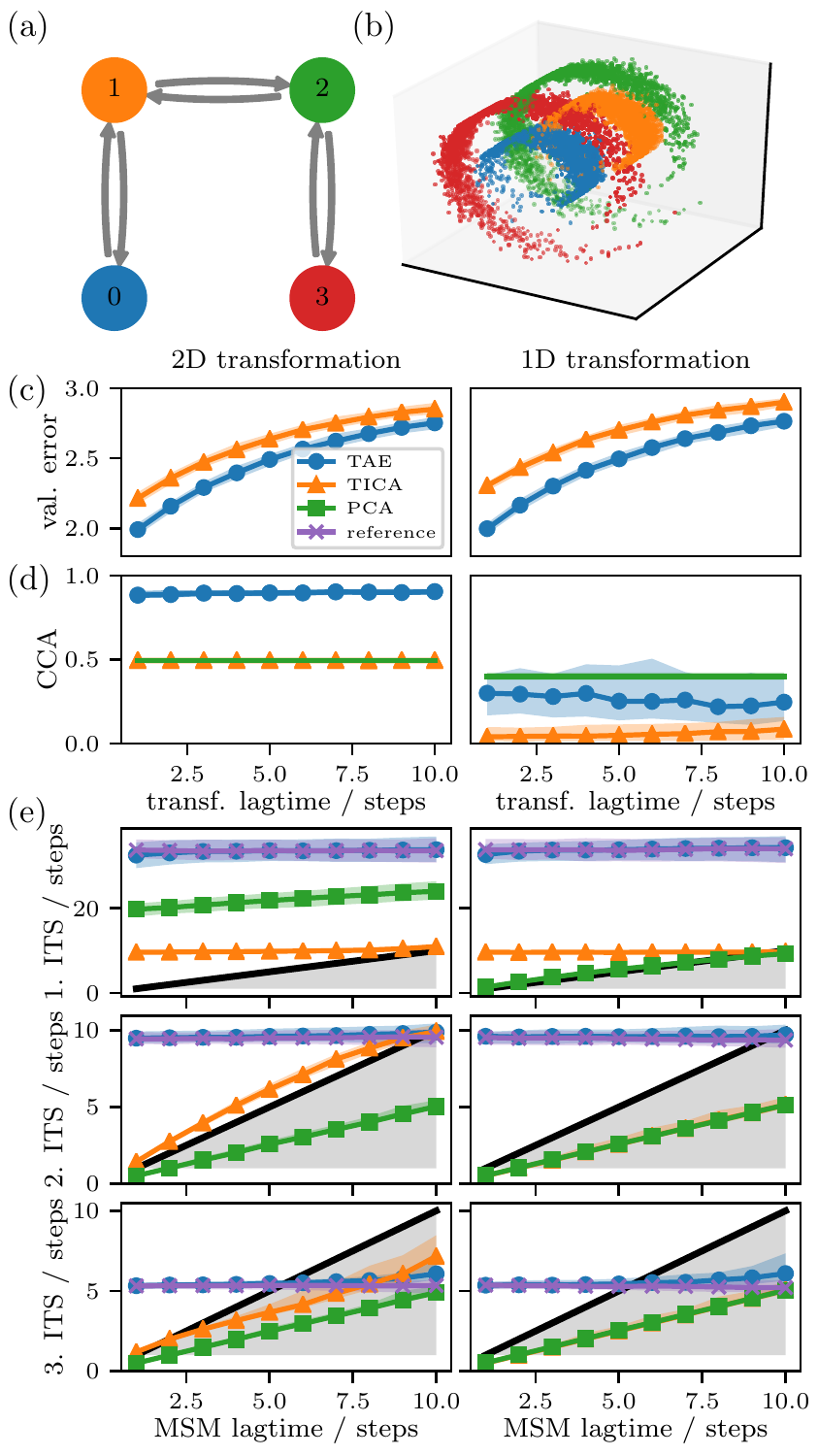}

\caption{\label{fig:swissroll}\textbf{Three-dimensional four-state system
that is not linearly separable}. (a) Network of the four-state HMM
toy model. (b) Emissions in a three-dimensional swissroll shape. (c)
Regression error of the validation set. (d) Canonical correlation
between the true HMM time series with the one- or two-dimensional
encoding, as a function of the lagtime $\tau$. (e) The first three
implied timescales (ITS) obtained from MSMs constructed on the encoding
space. The TAE and TICA were trained at a lagtime of one simulation
step. All lines show the median over 100 realizations, shaded areas
indicate 68\% percentiles.}
\end{figure}

The second toy model is based on a four-state hidden Markov model
(HMM), which emits isotropic Gaussian noise in the two-dimensional
$x/y$-plane, with the means of the states located as shown in Fig.
\ref{fig:swissroll}a. To create a nonlinearly separable system, we
perform the operation
\[
\left(\begin{array}{c}
x\\
y
\end{array}\right)\mapsto\left(\begin{array}{c}
x\cos(x)\\
y\\
x\sin(x)
\end{array}\right)
\]
which produces a picture that is reminiscent of the swiss roll commonly
used as a benchmark for nonlinear dimension reduction (Fig. \ref{fig:swissroll}b).
For this toy model, we examine two- and one-dimensional encodings.
The TAE with $d=2$ uses a single hidden layer with 100 units in the
encoder and decoder part, while the TAE with $d=1$ uses two hidden
layers with 200 and 100 units in the encoder part and 100 and 200
units in the decoder.

In both cases, the time-lagged autoencoder outperforms TICA in terms
of reconstruction error (Fig. \ref{fig:swissroll}c). Again, the difference
is larger for small transformation lagtimes. Indeed, the TAE encoding
is nearly perfectly correlated with the true hidden states time series
(Fig. \ref{fig:swissroll}d), while both TICA and PCA are significantly
worse and nearly identical to each other. In the one-dimensional case,
all methods fail at obtaining a high correlation, indicating that
this system is not perfectly separable with a single coordinate, even
if it is nonlinear. 

MSMs constructed on the encoded space also indicate that the TAE perfectly
recovers the reference timescales at all lagtimes (Fig. \ref{fig:swissroll}e)
\textendash{} surprisingly this is also true for the one-dimensional
embedding, despite the fact that this embedding is not well correlated
with the true hidden time series. MSMs build on either the TICA or
PCA space are systematically underestimated and mostly show no sign
of convergence.

\subsection{Molecular dynamics data of alanine dipeptide}

\begin{figure}
\includegraphics{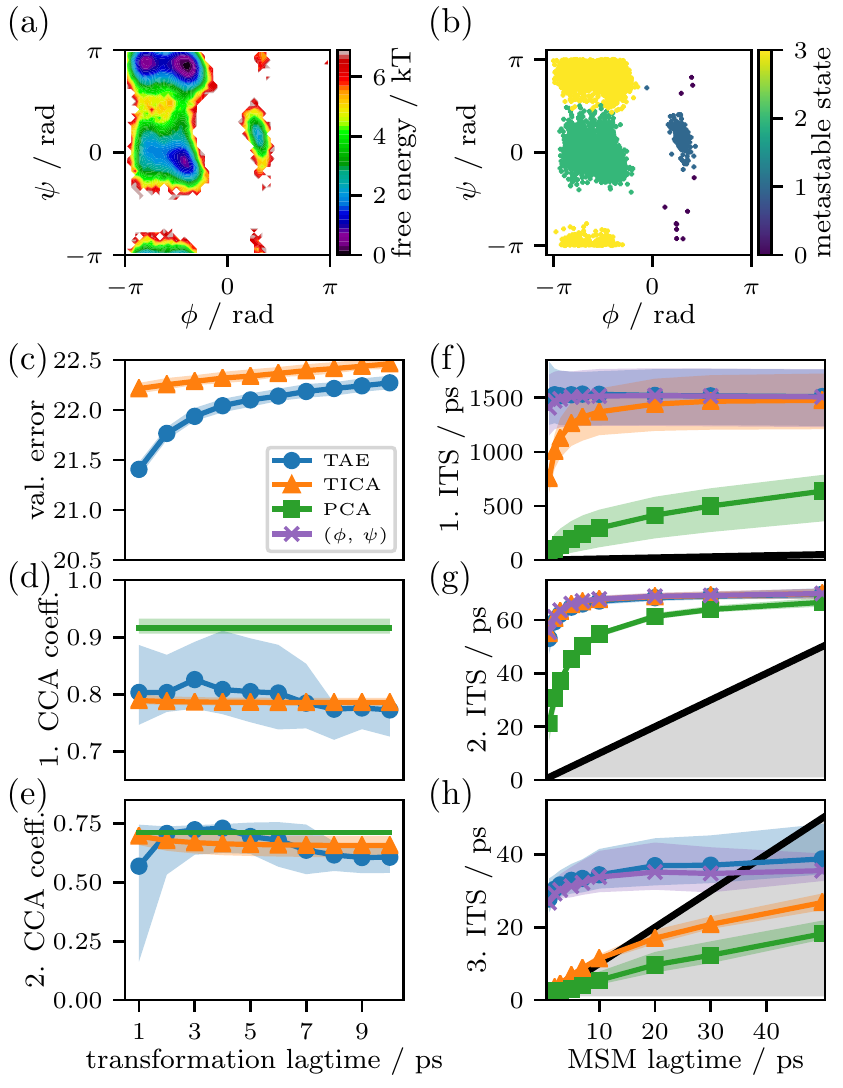}

\caption{\label{fig:ala2}\textbf{Alanine dipeptide kinetics estimated from
the time series in 30 heavy atom coordinates}. (a) Free energy surface
in the well-known space of $\phi/\psi$ backbone dihedrals. (b) Metastable
partition into the four most slowly-interconverting states. (c) Regression
error of the validation set. (d, e) Canonical correlations between
the $\phi/\psi$ dihedral representation and the two-dimensional encoded
spaces found by TAE, TICA or PCA. (f-g) First three implied timescales
(ITS) of MSMs constructed in the encoded space. The TAE was trained
at lagtime of 3ps and TICA at a lagtime 1ps. All lines show the median
over 100 bootstrapped samples, shaded areas indicate 68\% percentiles.}
\end{figure}

Our third example involves real MD data from three independent simulations
of 250 ns each \citep{oom-feliks,acemd} from which we repeatedly
bootstrap five sub-trajectories of length 100 ns. The features on
which we apply the encoding are RMSD-aligned heavy atom positions
which yield an $N=30$-dimensional input space. Although we do not
know the optimal two-dimensional representation, we assume that the
commonly used ($\phi$,$\psi$) backbone dihedrals contain all the
relevant long-time behavior of the system (except for methyl rotations
which do not affect the heavy atoms \citep{YiEtAl_JCP13_NeutronScatteringII}),
and we thus use these dihedral angles as a reference to compare our
encoding spaces to. Fig. \ref{fig:ala2}a and b show the free energy
surface for the reference representation and the assignment of ($\phi$,$\psi$)-points
to the four most slowly-interconverting metastable states. 

The TAE outperforms TICA in terms of the regression error (Fig. \ref{fig:ala2}c).
While all three methods find a two-dimensional space that correlates
relatively well with the ($\phi$,$\psi$)-plane, PCA achieves, surprisingly
the best correlation (Fig. \ref{fig:ala2}d-e), while the TAE and
TICA are similar. This result is put into perspective by the performances
of MSMs built upon the encoding space (Fig. \ref{fig:ala2}f-h). Here,
TAE clearly performs best. The TICA MSM does converge to the first
two relaxation timescales, although slower than the TAE in the first
relaxation timescale, while its convergence of the third relaxation
timescale is too slow to be practically useful. PCA performs poorly
for all relaxation timescales.

\section{Conclusion}

We have investigated the performance of a special type of deep neural
network, the time-lagged autoencoder, to the task of finding low-dimensional,
nonlinear embeddings of dynamical data. We have first shown that a
linear time-lagged autoencoder is equivalent to time-lagged canonical
correlation analysis, and for the special case of statistically time-reversible
data equivalent to the time-lagged independent component analysis
commonly used in the analysis of MD data. However, in many datasets,
the metastable states are not linearly separable and there is thus
no low-dimensional linear subspace that will resolve the slow processes,
resulting in large approximation errors of MSMs and other estimators
of kinetics or thermodynamics. In these cases, the traditional variational
approach puts the workload on the user who can mitigate this problem
by finding suitable feature transformations of the MD coordinates,
e.g., to contact maps, distances, angles or other other nonlinear
functions in which the metastable states may be linearly separable.
In a deep TAE, instead, we take the perspective that the nonlinear
feature transformation should be found automatically by an optimization
algorithm. Our results on toy models and MD data indicate that this
is indeed possible and low-dimensional representations can be found
that outperform those found by naive TICA and PCA.

Our approach is closely related to the previously proposed VAMPnet
approach that performs a simultaneous dimension reduction and MSM
estimation by employing the variational approach of Markov processes
\citep{MardtEtAl_VAMPnets}. By combining the theoretical results
from this paper with those of \citep{NoeNueske_MMS13_VariationalApproach,WuNoe_VAMP},
it is clear that in the linear case all these methods are equivalent
with TCCA, TICA, Koopman models or MSMs, depending on the type of
inputs used, and whether the data are reversible or nonreversible.
We believe that there is also a deeper mathematical relationship between
these methods in the nonlinear case, e.g., when deep neural networks
are employed to learn the feature transformation, but this relationship
is still elusive. Both the present approach, that minimizes the TAE
regression error in the input space, as well as the variational approach,
that maximizes a variational score in the feature space \citep{NoeNueske_MMS13_VariationalApproach,WuNoe_VAMP},
are suitable to conduct hyper-parameter search \citep{McGibbonPande_JCP15_CrossValidation}.
The present error model (\ref{eq:minimal_error}) is based on least
square regression, or in other words, on the assumption of additive
noise in the configuration space, while VAMPnets do not have this
restriction. Also, VAMPnets can incorporate the MSM estimation in
a single end-to-end learning framework. On the other hand, the autoencoder
approach has the advantage that, in addition to the feature encoding,
a feature decoding back to the full configuration space is learned,
too. Future studies will investigate the strengths and weaknesses
of both approaches in greater detail.\nocite{matplotlib}

\section*{Acknowledgments}

We are grateful for insightful discussions with Steve Brunton, Nathan
Kutz, Andreas Mardt, Luca Pasquali, and Simon Olsson. We gratefully
acknowledge funding by European Commission (ERC StG 307494 ``pcCell'')
and Deutsche Forschungsgemeinschaft (SFB 1114/A04). 

\bibliographystyle{aipnum4-1}
\bibliography{manuscript}

\end{document}